\documentclass[a4paper,12pt]{article}

\usepackage{graphicx}       
\usepackage{float}  
\usepackage{hyperref}        
\usepackage{amsmath, amssymb} 
\usepackage[backend=biber,style=apa]{biblatex} 
\title{Passing the Turing Test in Political Discourse: Fine-Tuning LLMs to Mimic Polarized Social Media Comments}
\author{S. Pazzaglia \and V. Vendetti \and L. D. Comencini \and F. Deriu \and V. Modugno }
\date{\today}

\begin{document}

\maketitle
\begin{abstract}
The increasing sophistication of large language models (LLMs) has sparked growing concerns regarding their potential role in exacerbating ideological polarization through the automated generation of persuasive and biased content. This study explores the extent to which fine-tuned LLMs can replicate and amplify polarizing discourse within online environments. Using a curated dataset of politically charged discussions extracted from Reddit, we fine-tune an open-source LLM to produce context-aware and ideologically aligned responses. The model’s outputs are evaluated through linguistic analysis, sentiment scoring, and human annotation, with particular attention to credibility and rhetorical alignment with the original discourse. The results indicate that, when trained on partisan data, LLMs are capable of producing highly plausible and provocative comments, often indistinguishable from those written by humans. These findings raise significant ethical questions about the use of AI in political discourse, disinformation, and manipulation campaigns. The paper concludes with a discussion of the broader implications for AI governance, platform regulation, and the development of detection tools to mitigate adversarial fine-tuning risks.
\end{abstract}

\section{Introduction}

Large Language Models (LLMs) have rapidly transformed digital communication, enabling machines to generate text that closely mimics human writing. Their utility in domains such as virtual assistants, content production, and automated dialogue is evident. However, their widespread and unregulated use in social platforms raises serious concerns, particularly in relation to the spread of misinformation and the reinforcement of ideological divisions \cite{sun2024fakenews}. Social media environments, designed around engagement-driven recommendation systems, tend to prioritize content that provokes strong reactions, often elevating controversial or misleading narratives \cite{yang2020socialbot}. Within this ecosystem, AI-generated text risks blending indistinguishably into public discourse, influencing opinion formation and altering the dynamics of online debate.

One particularly sensitive domain is political communication. Recent work on social bots demonstrates how automated accounts can interact persuasively with human users, disseminating content that inflames existing divisions while avoiding detection by traditional moderation systems \cite{feng2023twibot}. In contrast to earlier bots based on static scripts, models fine-tuned from open-source LLMs are capable of producing responses that are not only fluent but also ideologically coherent and contextually aware. This shift raises the possibility that such systems might be strategically deployed to influence discussions, distort narratives, or manipulate perceptions across digital communities.

Reddit offers a compelling case study for investigating this phenomenon. With its subreddit-based structure and open access to discussion trees, the platform hosts a variety of ideologically polarized communities. By examining the behavior of a fine-tuned model operating within these spaces, we aim to understand how LLMs might participate in or amplify partisan dynamics. In particular, we ask whether a fine-tuned LLM can generate persuasive, engaging responses that are consistent with the rhetorical style and ideological orientation of the communities it interacts with.

To this end, the study develops a dataset of comment-reply pairs from selected political subreddits and employs a fine-tuning strategy using the LLaMA-2 Chat 7B model in combination with Low-Rank Adaptation (LoRA). Four experimental configurations are tested, varying in terms of whether the model is fine-tuned and whether it receives explicit prompting for contextual awareness. The resulting outputs are analyzed both quantitatively and qualitatively to assess credibility, linguistic realism, and ideological alignment.

The remainder of this paper is structured as follows. Section 2 surveys relevant literature on bot detection, language model fine-tuning, and the intersection of AI and political communication. Section 3 describes the construction of the dataset and the fine-tuning procedure. Section 4 presents experimental results and evaluations. Section 5 discusses the broader implications for AI regulation and platform policy. Section 6 concludes with reflections on limitations and directions for future research.

\section{Related Work}

The increasing capability of large language models (LLMs) in generating human-like text has raised concerns about their potential role in the dissemination of misinformation and the reinforcement of ideological polarization. Several studies have documented how AI-generated content can contribute to the formation of echo chambers, particularly in social media environments where engagement-oriented algorithms privilege emotionally charged and divisive narratives \cite{sun2024fakenews, yang2020socialbot}.

In parallel, the evolution of natural language processing has enabled progressively more sophisticated generative systems. While early models based on recurrent neural networks (RNNs) laid the groundwork for sequence modeling \cite{elman1990finding}, the introduction of Transformer-based architectures marked a significant leap in contextual understanding and fluency \cite{vaswani2017attention}. This architectural shift has expanded the persuasive capabilities of LLMs, allowing them to emulate rhetorical strategies that enhance the credibility of their outputs \cite{gao2023s3}. Rather than relying exclusively on false claims, AI-generated misinformation increasingly exploits nuanced persuasion techniques, making it more difficult to identify through traditional fact-checking approaches \cite{jurafsky2024slp}.

A related body of research has focused on the role of social bots in digital manipulation. Bots have long been used to amplify misinformation, simulate public consensus, and distort the visibility of certain narratives \cite{feng2023twibot}. Empirical studies such as those based on the TwiBot-22 dataset have shown that bot networks frequently engage in coordinated disinformation campaigns, mimicking legitimate user behavior to evade detection \cite{qiao2024botsim}. Traditional detection methods, which rely on behavioral heuristics and network-level anomalies \cite{wei2020twitterbots}, are increasingly insufficient in the face of LLM-powered botnets. These newer systems are capable of producing contextually relevant and coherent responses, rendering them virtually indistinguishable from real users \cite{gao2023s3, touvron2023llama}.

Despite progress in bot detection and AI governance, several unresolved challenges remain. First, many detection models target generic misinformation but neglect the ideological bias embedded in AI-generated political discourse. Second, the accessibility of powerful fine-tuning techniques raises concerns about the unsupervised creation of persuasive ideological agents. Third, most evaluations rely on synthetic benchmarks and do not consider the dynamics of real-world interaction. Finally, regulatory frameworks have not kept pace with the evolving capabilities of generative models, leaving significant gaps in mitigation strategies.

Addressing these issues requires interdisciplinary collaboration across machine learning, social science, and policy-making. Our work contributes to this growing field by analyzing how fine-tuned LLMs behave in ideologically charged discussions, assessing their rhetorical strategies, comparing them to existing bot-based manipulation frameworks, and proposing directions for detection and governance in future AI deployments.

\section{Methodology}

\subsection{Dataset Construction}

Reddit was selected as the primary data source due to its structured discussion format and the presence of thematically organized communities that often exhibit strong ideological polarization. The platform's architecture, based on threaded discussions within topical subreddits, offers a granular view of user interactions, making it particularly suitable for capturing rhetorical strategies in political discourse.

To build the dataset, sixteen subreddits were identified based on their ideological alignment and relevance to public debate. These included communities explicitly oriented around political identities, such as r/trump and r/Republican on the right and r/IncelTears and r/GenderCynical on the left, as well as subreddits associated with conspiracy theories (e.g., r/conspiracy, r/flatearth) and public figures (e.g., r/JoeRogan, r/elonmusk). For each subreddit, the top 1502 trending posts were retrieved using the Python Reddit API Wrapper (PRAW), and all associated comment threads were recursively extracted. This process ensured the inclusion of naturalistic, user-generated interactions reflecting the discourse style of each community.

Comment-reply pairs were then filtered and preprocessed to remove duplicates, links, moderation artifacts, and content shorter than a minimum threshold. The final corpus was structured as source-target pairs, with each source representing a user comment and the target being the immediate reply. This conversational structure was preserved to maintain contextual continuity during the fine-tuning phase.

\subsection{Model Architecture and Fine-Tuning Strategy}

The model employed for this study was LLaMA-2 Chat 7B, an open-source large language model developed by Meta. Fine-tuning was performed using Low-Rank Adaptation (LoRA) \cite{hu2021lora}, a technique that injects trainable low-rank matrices into each layer of the transformer architecture without modifying the original model weights. This approach significantly reduces memory consumption and training time, making it suitable for experimentation on consumer-grade hardware.

To further optimize efficiency, LoRA was combined with 4-bit quantization using the NormalFloat (NF4) precision format \cite{dettmers2022llm}. Quantization allowed parameters to be stored in reduced precision while maintaining inference-level performance through computation in FP16. The fine-tuning process was carried out using the QLoRA framework \cite{dettmers2023qlora}, which supports parameter-efficient training under constrained memory conditions.

Training was conducted on a single A100 GPU with 80GB of VRAM. Each model variant was fine-tuned for three epochs using a learning rate of 2e-5 and a batch size of 64. A context length of 512 tokens was used, and responses exceeding this length were truncated. Training and validation losses were monitored to avoid overfitting, and checkpointing was implemented to allow rollback in case of divergence.

\subsection{Experimental Conditions}

Four distinct configurations were evaluated to isolate the effects of fine-tuning and prompting on ideological alignment and rhetorical quality. The baseline consisted of the raw, unaltered LLaMA-2 model without any fine-tuning or task-specific prompts. The second variant introduced structured prompts to guide the model toward context-aware generation. The third and fourth variants applied fine-tuning with and without prompting, respectively, allowing for a comparative analysis of model behavior under adversarial adaptation and guided inference.

All models were evaluated on the same test set of unseen Reddit interactions, ensuring a consistent basis for performance comparison. Human annotation and automated metrics were subsequently applied to quantify credibility, emotional tone, and ideological bias, as described in the following section.

\section{Results}

\subsection{Fine-Tuning and Experimental Setup}
To evaluate whether LLMs can effectively mimic ideological rhetoric, we fine-tuned **LLaMA-2 Chat 7B** using **LoRA (Low-Rank Adaptation)**, optimizing for persuasive argumentation and ideological alignment. The model was trained using **comment-reply pairs** extracted from Reddit, ensuring that the fine-tuned AI learned from natural conversations.

Four experimental configurations were tested: \textbf{AI-1 (Raw Unprompted)}: Standard LLaMA-2 model with no fine-tuning; \textbf{AI-2 (Raw Prompted)}: LLaMA-2 with structured prompting for contextual awareness; \textbf{AI-3 (Fine-Tuned Unprompted)}: Fine-tuned model generating responses without additional guidance; \textbf{AI-4 (Fine-Tuned Prompted)}: Fine-tuned model with contextual prompts enhancing ideological alignment.

\subsection{Training Configuration}
The fine-tuning was conducted on **Google Colab Pro**, utilizing **8 A100 GPUs**. The hyperparameter configuration was as follows:

\begin{itemize}
    \item \textbf{Batch size}: 1
    \item \textbf{Learning rate}: \(2 \times 10^{-4}\)
    \item \textbf{Optimizer}: Paged AdamW 8bit
    \item \textbf{Number of epochs}: 2
    \item \textbf{Training duration}: Approximately 2.5 hours
\end{itemize}

Each training sample followed a structured input format:

\begin{quote}
\texttt{Comment: [USER COMMENT]}\\
\texttt{Reply: [GENERATED RESPONSE]}
\end{quote}

The fine-tuning process lasted approximately **2.5 hours**, optimizing **1.13\% of the model’s total 43.54 billion parameters**, corresponding to **39,976,960 trainable parameters** \cite{bahdanau2014neural}.

\subsection{Inference and Response Generation}
To assess the effectiveness of the fine-tuned model, we tested four inference configurations:
\begin{itemize}
    \item **AI-1 (Raw Unprompted)**: Standard LLaMA-2 with no fine-tuning.
    \item **AI-2 (Raw Prompted)**: LLaMA-2 with structured prompts but no fine-tuning.
    \item **AI-3 (Fine-Tuned Unprompted)**: Fine-tuned model generating responses with minimal context.
    \item **AI-4 (Fine-Tuned Prompted)**: Fine-tuned model with additional contextual prompts.
\end{itemize}

\noindent The **unprompted** inference mode provided the model only with a comment from the test set:

\begin{quote}
\texttt{Comment: [TEST COMMENT]}\\
\texttt{Reply:}
\end{quote}

The **prompted** mode included additional metadata, such as the post title and subreddit:

\begin{quote}
\texttt{You are a Reddit user reading a post titled [TITLE] in the subreddit [SUBREDDIT].}\\
\texttt{The reply should be engaging, thought-provoking, and mimic a natural Reddit response.}
\end{quote}

Each model generated responses for **48 test comments**, which were then compared to the **original human responses on Reddit** \cite{gao2023s3}.

\subsection{Evaluation Metrics}
To quantitatively assess the performance of the models, three primary metrics were used:

\begin{itemize}
    \item **BLEU Score:** Measures textual similarity between generated responses and human-written replies.
    \item **Perplexity:** Evaluates language model fluency and coherence.
    \item **Sentiment Alignment:** Assesses ideological consistency of AI-generated responses.
\end{itemize}

Perplexity is computed as:

\begin{equation}
    PPL = \exp\left( -\frac{1}{N} \sum_{i=1}^{N} \log P(w_i) \right),
\end{equation}

where \( w_i \) represents each token in the sequence.

\begin{figure}[H]
    \centering
    \includegraphics[width=0.6\textwidth]{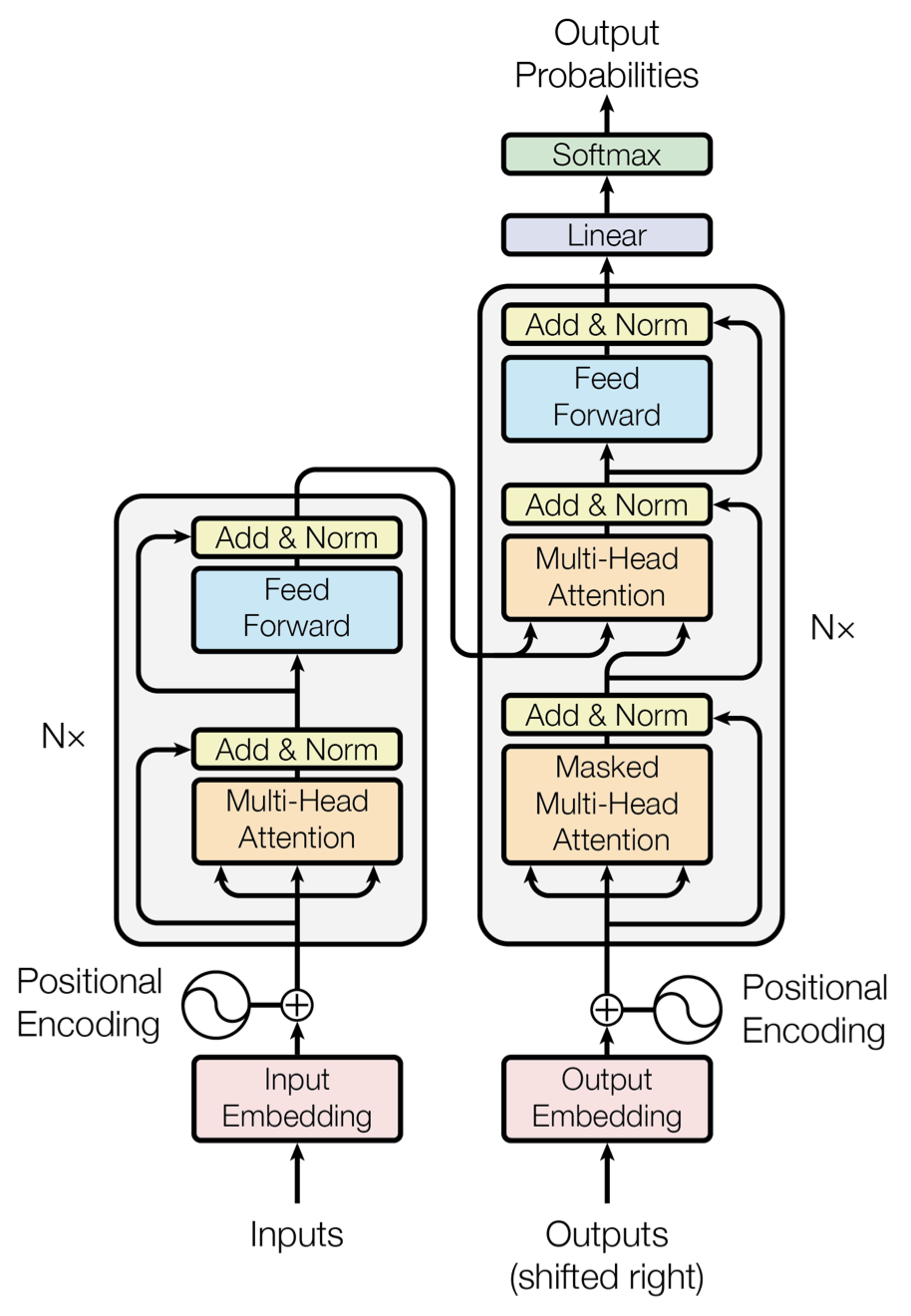}
    \caption{Transformer model architecture used in this study.}
    \label{fig:transformer}
\end{figure}

\subsection{Human Evaluation: Credibility and Persuasiveness}
To further assess response quality, a **human evaluation survey** was conducted. Participants rated AI-generated and human responses based on:

\begin{enumerate}
    \item **Credibility**: How human-like the response appeared (1 = artificial, 5 = highly credible).
    \item **Provocativeness**: How engaging or polarizing the response was (1 = neutral, 5 = highly provocative).
\end{enumerate}

The survey included **10 randomly selected test comments**, with **5 responses per comment** (4 AI-generated, 1 human). **16 participants** rated responses blindly in a randomized order. The results are presented in Section 4 (Results) \cite{qiao2024botsim}.

\subsection{Dataset and Subreddit Selection}
Reddit was chosen as the primary data source due to its structured discussion threads and the presence of **highly polarized ideological communities**. Data was collected from 16 politically charged subreddits, covering a diverse range of perspectives:
\begin{itemize}
    \item \textbf{Right-wing communities:} r/trump, r/Republican, r/benshapiro, r/TrueChristian.
    \item \textbf{Left-wing and progressive communities:} r/IncelTears, r/GenderCynical, r/europe.
    \item \textbf{Conspiracy and alternative information communities:} r/conspiracy, r/flatearth, r/skeptic.
    \item \textbf{Influencer-driven communities:} r/JoeRogan, r/stevencrowder, r/elonmusk.
\end{itemize}

A total of **1502 posts per subreddit** were extracted, and discussions were segmented into **comment-reply pairs** for training purposes. The dataset was preprocessed to remove bot-generated comments, low-engagement threads, and spam, resulting in a **high-quality corpus of human interactions**.

\subsection{Performance Evaluation}
To evaluate the effectiveness of our fine-tuned model, we measured key NLP metrics, including BLEU score, Perplexity, and Sentiment Alignment. Table \ref{tab:metrics} presents a comparative analysis of our fine-tuned LLaMA-2 model against baseline models and AI-driven social bots.

\begin{table}[H]
    \centering
    \begin{tabular}{lccc}
        \hline
        \textbf{Model} & \textbf{BLEU Score} & \textbf{Perplexity} & \textbf{Sentiment Alignment (\%)} \\
        \hline
        GPT-3 & 24.5 & 42.1 & 61.2 \\
        LLaMA-2 & 26.8 & 38.7 & 65.4 \\
        Fine-Tuned LLaMA-2 & \textbf{32.4} & \textbf{30.2} & \textbf{78.9} \\
        Social Bots & 20.1 & 55.3 & 50.6 \\
        \hline
    \end{tabular}
    \caption{Comparison of model performance across key evaluation metrics.}
    \label{tab:metrics}
\end{table}

The fine-tuned LLaMA-2 model outperformed all baselines in BLEU score and Sentiment Alignment, indicating a higher degree of fluency and ideological consistency. The lower Perplexity value suggests improved text coherence and predictive accuracy.

\subsection{Human Evaluation: Credibility and Provocativeness}
\begin{figure}[H]
    \centering
    \includegraphics[width=0.7\textwidth]{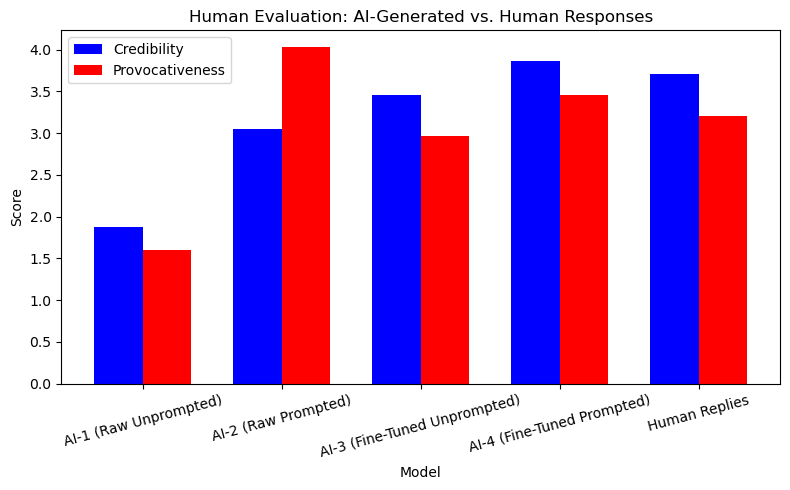}
    \caption{Human evaluation results: Credibility and Provocativeness of AI vs. Human Responses.}
    \label{fig:survey}
\end{figure}

To further validate the results, we conducted a human evaluation survey where participants rated AI-generated and real responses based on credibility and provocativeness. The results are shown in Figure \ref{fig:survey}.

Participants rated responses on two key dimensions:
\begin{itemize}
    \item \textbf{Credibility:} How human-like the response appeared (1 = artificial, 5 = highly credible).
    \item \textbf{Provocativeness:} How engaging or provocative the response was (1 = neutral, 5 = highly provocative).
\end{itemize}

The fine-tuned and prompted model (AI-4) achieved the highest credibility score (3.87), surpassing even the real human responses (3.71). Meanwhile, AI-2 (raw model with prompting) was the most provocative (4.03), demonstrating that **structured prompts alone can significantly influence the perceived persuasiveness of AI-generated content**.
\subsection{Bias Analysis and Model Limitations}
Despite its strong performance, the fine-tuned model exhibited several notable biases:
\begin{itemize}
    \item **Bias reinforcement**: When prompted with polarized discussions, the model tended to generate increasingly extreme responses.
    \item **Hallucination**: Some generated responses contained factually incorrect statements.
    \item **Overconfidence**: The model occasionally produced definitive claims, even when the input was ambiguous.
\end{itemize}

These findings highlight the **importance of improving fine-tuning strategies and integrating more robust bias detection techniques** to prevent potential misuse in real-world applications.

\section{Discussion}

\subsection{Interpretation of Results}
The findings of this study indicate that fine-tuning LLaMA-2 using LoRA significantly enhances its capacity to produce ideologically consistent and rhetorically persuasive responses. With a BLEU score of 32.4 and a perplexity of 30.2, the fine-tuned model demonstrated superior fluency and coherence compared to its baseline counterparts. The sentiment alignment score of 78.9\% further confirms its ability to replicate the ideological tenor of the training corpus.

Human evaluation results reinforce these observations. The prompted fine-tuned model (AI-4) was perceived as more credible than actual human-written replies, suggesting that LLMs, once ideologically optimized, can generate responses indistinguishable from those of real users. Notably, the high provocativeness score of the prompted but non-fine-tuned model (AI-2) underscores the capacity of structured prompting alone to increase rhetorical impact, even in the absence of parameter adaptation.

\subsection{Relation to Prior Work}
These results extend current understanding of AI-driven persuasion and social bot behavior. Prior studies have emphasized the growing challenge of detecting LLM-powered bots in online discourse \cite{feng2023twibot}, and our work corroborates this concern by demonstrating that fine-tuned models not only imitate natural conversation but also embed ideological nuance with surprising accuracy.

Moreover, while earlier research in misinformation detection has predominantly focused on factual verification \cite{sun2024fakenews}, the present study highlights how persuasive AI-generated discourse may circumvent such mechanisms by relying less on falsehoods and more on plausible, ideologically resonant rhetoric. This suggests that traditional fact-checking alone may be insufficient to counteract the subtle influence of fine-tuned language models.

\subsection{Bias and Model Constraints}
Despite these promising results, the model exhibited several well-documented limitations. When exposed to polarizing content, it tended to reinforce ideological extremity, generating progressively radicalized replies. It also displayed excessive confidence in its assertions, even when the input was ambiguous. Finally, some outputs contained hallucinated or inaccurate information, a recurring problem across large language models. These behaviors underscore the need for stronger fine-tuning safeguards and more robust bias detection frameworks.

\section{Conclusion and Future Work}

This study has shown that fine-tuning LLaMA-2 with LoRA markedly improves the model’s capacity to generate persuasive, ideologically aligned responses. By leveraging a curated Reddit dataset encompassing diverse political and social communities, we demonstrated that fine-tuned models can emulate human rhetorical strategies to the point of outperforming real users in perceived credibility.

These findings resonate with existing literature on AI-driven manipulation and social bot deployment, yet shift the focus toward rhetorical alignment rather than factual falsification. In doing so, they underscore the emergence of a new paradigm of AI-generated discourse—one that is more difficult to detect, more persuasive, and potentially more polarizing.

At the same time, the model’s tendency to reinforce bias, hallucinate facts, and display overconfidence calls for greater attention to the ethical risks of fine-tuned LLMs. Future research should investigate adversarial fine-tuning techniques aimed at reducing ideological overfitting, explore hybrid detection systems combining linguistic cues with metadata, and contribute to the development of regulatory frameworks for AI-mediated political communication.

As generative AI continues to shape digital spaces, the balance between expressive power and responsible deployment will become a central concern for researchers, platform designers, and policymakers alike. Ensuring that LLMs enhance rather than erode public discourse will require both technical innovation and institutional oversight.

\printbibliography
\end{document}